\documentclass[aip,cha,amsmath,amssymb,reprint,author-year,author-numerical]{revtex4-1}
\usepackage{epsfig}
\usepackage{color}
\usepackage{graphicx}
\usepackage{dcolumn}
\usepackage{bm}
\input epsf

\usepackage[normalem]{ulem} 

\graphicspath{{figures/}}

\begin{document}


\title{Detecting disturbances in network-coupled dynamical systems with machine learning}

\author{Per Sebastian Skardal}
\email{persebastian.skardal@trincoll.edu}
\affiliation{Department of Mathematics, Trinity College, Hartford, CT 06106, USA}

\author{Juan G. Restrepo}
\affiliation{Department of Applied Mathematics, University of Colorado at Boulder, Boulder, CO 80309, USA}

\begin{abstract}
Identifying disturbances in network-coupled dynamical systems without knowledge of the disturbances or underlying dynamics is a problem with a wide range of applications. For example, one might want to know which nodes in the network are being disturbed and identify the type of disturbance. Here we present a model-free method based on machine learning to identify such unknown disturbances based only on prior observations of the system when forced by a known training function. We find that this method is able to identify the locations and properties of many different types of unknown disturbances using a variety of known forcing functions. We illustrate our results both with linear and nonlinear disturbances using food web and neuronal activity models. Finally, we discuss how to scale our method to large networks.
\end{abstract}


\maketitle

\begin{quotation}
Despite a wide range of potential applications, identifying disturbances made to network-coupled dynamical systems is a difficult problem due to the complex nature of interactions between different units. Even a disturbance made to a single node can be challenging to localize due to the propagation of behavior through the network. Here we present a model-free method using machine learning techniques, specifically reservoir computing, to identify disturbances to network-coupled dynamical systems. This method assumes no knowledge of the underlying dynamics nor the disturbance itself. All that is needed is sufficient observations of the system under the influence of a known forcing function. We show using examples from ecology and neuroscience that the method robustly identifies the disturbances made to the system for a relatively simple set of forcing functions used for training. Moreover, we show that the method is scalable to large networks using a pseudo-parallelization architecture. 
\end{quotation}

\section{Introduction}\label{sec:01}

Machine learning techniques have proven to be extremely useful for data-driven modeling and prediction of complex systems \cite{Tanaka2019,Brunton2019,Tang2020,Nakajima2021}. In many of these applications, a machine learning system is trained to learn and replicate the dynamics of a nonlinear system from noisy or partially observed data, and then the machine learning system is used, for example, to forecast the dynamics \cite{Pathak2018PRL,SrinivasanPRL2022,Shahi2021Frontiers,Shahi2022MLA}, to estimate Lyapunov exponents  \cite{Pathak2017Chaos} or unstable periodic orbits \cite{Zhu2019}, to infer network coupling \cite{Banerjee2019Chaos}, or to predict extreme events \cite{Pyragas2020PLA} and crises in non-stationary dynamical systems \cite{Kong2023Chaos,Patel2023Chaos}. 

Machine learning techniques can also be used without the need to replicate the intrinsic  dynamics of the system. For example, Ref.~\cite{Canaday2021JPhys} uses reservoir computers, a particular class of machine learning systems suited to modeling time-dependent systems, to learn the response of dynamical systems to stimuli and then design data-driven control algorithms. In Ref.~\cite{RestrepoArxiv}, we recently proposed a similar scheme to identify and suppress unknown disturbances to dynamical systems.   Detecting disturbances to nonlinear dynamical systems is a crucial problem with a wide range of applications, e.g., in engineering, and in particular in power grid networks~\cite{Upadhyaya2015IEEE,Mathew2016IEEE,Chen2015IEEE,Ferreira2016EPSR,Wang2019JISA,Delabays2021NJP,Delabays2022IFAC}, ecology~\cite{Meurant2012,Battisti2016}, fluid dynamics~\cite{Bewley1998JFM, Bewley2000PhysD}, and climate change~\cite{Verbesselt2010RSE}. In this paper, we extend our results to the identification of disturbances in network-coupled dynamical systems. Network-coupled systems present a particular challenge in identifying disturbances, as the ripple of an external force complicates the inference of both the location and nature of the disturbance~\cite{Nudell2013IEEE,Lee2018IEEE,Delabays2021NJP}. By extending the method presented in \cite{RestrepoArxiv} to network-coupled dynamical systems, we are able to robustly identify which nodes are disturbed, and the time course of the disturbances. We illustrate our approach with two examples: a food web and a system of excitatory and inhibitory neuron populations. The dynamics in these examples include both stationary and oscillatory dynamics, and linear and nonlinear disturbances. We also address the issue of the scalability of our approach to large networks. We believe our approach could be useful in applications where identifying the location of disturbances in networked dynamical systems is important, such as, for example, power grid systems.

The remainder of this paper is organized as follows. In Sec.~\ref{sec:02} we present the problem statement and describe the setup of the reservoir computer. In Sec.~\ref{sec:03} we present our first example: linear disturbances to a food web. In Sec.~\ref{sec:04} we present our second example: nonlinear disturbances to a network of excitatory and inhibitory neuron populations. In Sec.~\ref{sec:05} we explore the scalability of our approach to larger networks using an ensemble of node-specific reservoirs. In Sec.~\ref{sec:06} we conclude with a discussion of our results.

\section{Problem Statement and Setup of the Reservoir Computer}\label{sec:02}

We consider here systems of $N$ network-coupled dynamical systems whose states $\bm{x}_i(t)\in\mathbb{R}^D$ for $i=1,\dots,N$ are governed by a system of differential equations of the form
\begin{align}
\dot{\bm{x}}_i = \bm{F}_i(\bm{x}_i,\bm{X},\bm{g}_i(t))\label{eq:01}
\end{align}
where $\bm{X}(t)=[\bm{x}_1(t)^T,\dots,\bm{x}_N(t)^T]^T\in\mathbb{R}^{ND}$ is the collection of all state vectors $\bm{x}_i(t)$ organized in a single vector, $\bm{g}_i(t)\in\mathbb{R}^D$ is the disturbance to the state at node $i$, and the vector field $\bm{F}_i:\mathbb{R}^D\times\mathbb{R}^{ND}\times\mathbb{R}^D\to\mathbb{R}^D$ incorporates the local intrinsic dynamics of state $i$, the effect of interactions with other states via $\bm{X}$ and some underlying network structure, and the local disturbance $\bm{g}_i$. In particular, we write the vector field $\bm{F}_i$ such that $\bm{F}_i(\bm{x}_i,\bm{X},\bm{0})=\bm{F}_{i,\text{intrinsic}}(\bm{x}_i,\bm{X})$ gives the intrinsic, i.e., undisturbed, system dynamics. In general, the disturbances $\bm{g}_i(t)$ may be linear, in which case the dynamics may be written $\bm{F}_i(\bm{x}_i,\bm{X},\bm{g}_i(t))=\bm{F}_{i,\text{intrinsic}}(\bm{x}_i,\bm{X})+\bm{g}_i(t)$, or nonlinear, in which case no such notational simplification can be made. 
Our goal is to develop a method by which the disturbances can be accurately identified in such a way that we can infer (i) precisely which element(s) of the system are disturbed and (ii) what the disturbance functions are. We assume that the system can be forced with a set of known {\it training forcing functions} $\bm{h}_i(t)\in\mathbb{R}^D$ for $i=1,\dots,N$, as
\begin{align}
\dot{\hat{\bm{x}}}_i = \bm{F}_i(\hat{\bm{x}}_i,\hat{\bm{X}},\bm{h}_i(t)),\label{eq:02}
\end{align}
and each $\hat{\bm{x}}_i(t)$ can be observed for sufficiently long time. (From now on, a hat will indicate quantities during the training phase). Then, a machine learning system is trained to approximate each $\bm{h}_i(t)$ given all $\hat{\bm{x}}_i(t)$. The trained machine learning system is then used to infer $\bm{g}_i(t)$ from observations of $\bm{x}_i(t)$ obtained from Eq.~(\ref{eq:01}). Note that no knowledge of the intrinsic dynamics $\bm{F}_{i,\text{intrinsic}}$ or the disturbance functions $\bm{g}_i$ is required. 

\begin{figure}[t]
\centering
\epsfig{file =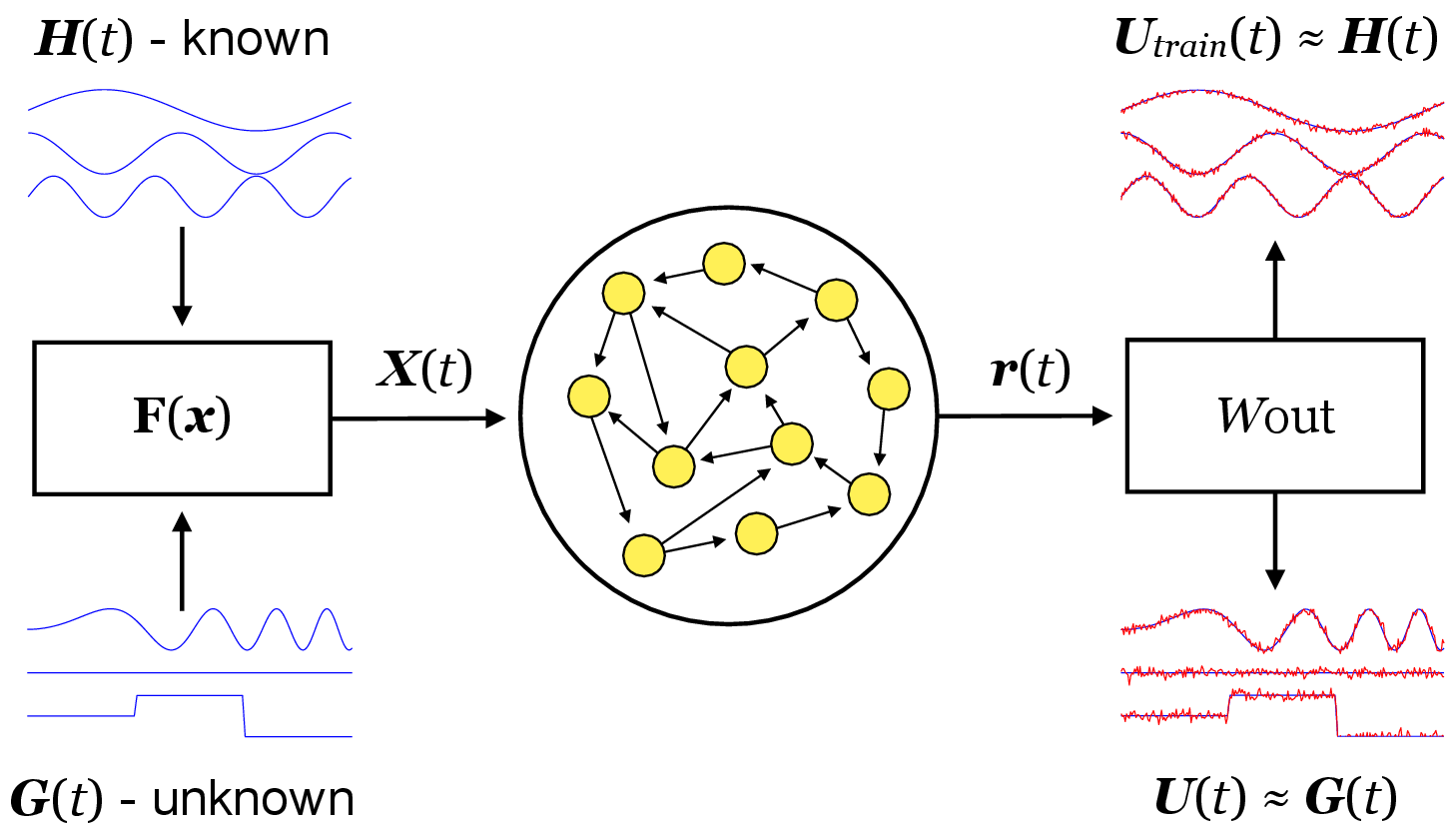, clip =,width=1.0\linewidth }
\caption{{\it Reservoir computer architecture.} Illustration of the setup of our reservoir computer. During the training phase the known forcing functions $\bm{h}_i(t)$ (organized in $\bm{H}(t)$) force the dynamics (top left), which produces the states $\hat{\bm{x}}_i(t)$ (organized in $\hat{\bm{X}}(t)$) and subsequently feed the reservoir (center), whose variables $\bm{r}(t)$ are used to train the output matrix $W_{\text{out}}$ to recover $\bm{U}(t)\approx\bm{H}(t)$ (top right). Next the system is disturbed with the unknown functions $\bm{g}_i(t)$ (organized in $\bm{G}(t)$) (bottom left), which produces $\bm{X}(t)$, then feeds the reservoir (center), at which point the new variables $\bm{r}(t)$ are used to recover $\bm{U}(t)\approx\bm{G}(t)$ (bottom right)}\label{fig1}
\end{figure}

As our machine-learning system implementation we will use reservoir computers, a class of machine learning systems particularly well-suited for time-dependent problems~\cite{Nakajima2021}. We assume that the training system given in Eq.~(\ref{eq:02}) [i.e., with known forcing functions $\bm{h}_i(t)$] is first run on a training interval $[-\hat{T},0]$, and a time-series of the observed state vector $\{\hat{\bm{X}}(-\hat{T}),\hat{\bm{X}}(-\hat{T} +\Delta t),\dots,\hat{\bm{X}}(0)\}$ is collected. At each time step, these variables are fed into the reservoir, a high-dimensional dynamical system with internal variables $\bm{r}\in \mathbb{R}^M$, where $M$ is the size of the reservoir. Here we implement the reservoir as
\begin{align}
\bm{r}(t+\Delta t) = \tanh[A \bm{r}(t) +W_{\text{in}} \hat{\bm{X}}(t) + 1],\label{eq:03}
\end{align}
where the $M\times M$ matrix $A$ is a sparse matrix representing the internal structure of the reservoir and the $M\times ND$ matrix $W_{\text{in}}$ is a fixed input matrix. At each time, the reservoir output $\bm{u}$ is constructed from the internal states as $\bm{U} = W_{\text{out}} \bm{r}$, where $\bm{U}(-n\Delta t)=[\bm{u}_1(t)^T,\dots,\bm{u}_N(t)^T]^T\in\mathbb{R}^{ND}$ organizes the vectors $\bm{u}_i(t)$ into a single vector and the $ND\times M$ output matrix $W_{\text{out}}$ is chosen so that the reservoir outputs ${\bf u}_i(t)$ are a good approximation to each known training forcing function $\bm{h}_i(t)$. Thus, the output matrix $W_{\text{out}}$ is the only component of the reservoir that needs to be trained, and this can be done by minimizing the cost function
\begin{align}
\sum_{n=0}^{\hat{T}/\Delta t} \| \bm{H}(-n\Delta t) - \bm{U}(-n\Delta t) \|^2 + \lambda \text{Tr}({W_{\text{out}} W_{\text{out}}^\text{T}}),\label{eq:04}
\end{align}
via a ridge regression procedure, where a small constant $\lambda \ge 0$ is used to prevent over-fitting and $\bm{H}(t)=[\bm{h}_1(t)^T,\dots,\bm{h}_N(t)^T]^T\in\mathbb{R}^{ND}$ organizes the vectors $\bm{h}_i(t)$ into a single vector. This procedure trains the reservoir to identify the forcing functions $\bm{h}_i(t)$ given the observations of $\hat{\bm{x}}_i(t)$, and subsequently the reservoir can be presented with a time series of the observed variables $\bm{x}_i(t)$ from Eq.~(\ref{eq:01}) in an interval $[0,T]$ and evolved as
\begin{align}
\bm{r}(t+\Delta t) = \tanh[A \bm{r}(t) +W_{\text{in}} \bm{X}(t)+1].\label{eq:05}
\end{align}
If the method works as intended, the reservoir output $\bm{U}(t) = W_{\text{out}} \bm{r}$ on the interval $[0,T]$ will be a good approximation to the unknown disturbance, $\bm{u}_i \approx \bm{g}_i$, where $\bm{g}_i(t)=\bm{0}$ for nodes $i$ that are not directly disturbed. We define the vector $\bm{G}(t)=[\bm{g}_1(t)^T,\dots,\bm{g}_N(t)^T]^T\in\mathbb{R}^{ND}$ to organize the disturbance vectors $\bm{g}_i(t)$ into a single large vector. Ref.~\cite{RestrepoArxiv} illustrates how the method performs when the intrinsic dynamics consists of a single Lorenz chaotic system.

In Fig.~\ref{fig1} we illustrate the setup of our reservoir computer. First, during the training phase, i.e., in the interval $[-\hat{T},0]$, the known forcing functions $\bm{h}_i(t)$ [organized in $\bm{H}(t)$] are fed into the reservoir (top left), while the observed system states $\hat{\bm{x}}_i(t)$ [organized in $\hat{\bm{X}}(t)$] force the reservoir dynamics (center), and the output matrix $W_{\text{out}}$ is trained to recover $\bm{U}(t)\approx\bm{H}(t)$ (top right). When training is complete, the unknown disturbances $\bm{g}_i(t)$ [organized in $\bm{G}(t)$] are fed into the reservoir (bottom left), while the new observed system states $\bm{x}_i(t)$ force the reservoir dynamics (center), and $\bm{U}(t)\approx\bm{G}(t)$ is recovered (bottom right).

Throughout this paper, unless otherwise specified, the reservoir matrix $A$ is a random matrix of size $M = 1000$ with entries $A_{ij}$ uniformly distributed in $[-0.5,0.5]$ with probability $6/M$ and otherwise $A_{ij}=0$, then rescaled to set its spectral radius to $1.2$. The input matrix $W_{\text{in}}$ is a random matrix where each entry is uniformly distributed in $[-0.01,0.01]$. The ridge regression regularization constant is $\lambda = 10^{-6}$. Training times $\hat{T}$ and time steps $\Delta t$ vary depending on the chosen dynamics and will be given below with each example.

\begin{figure}[t]
\centering
\epsfig{file =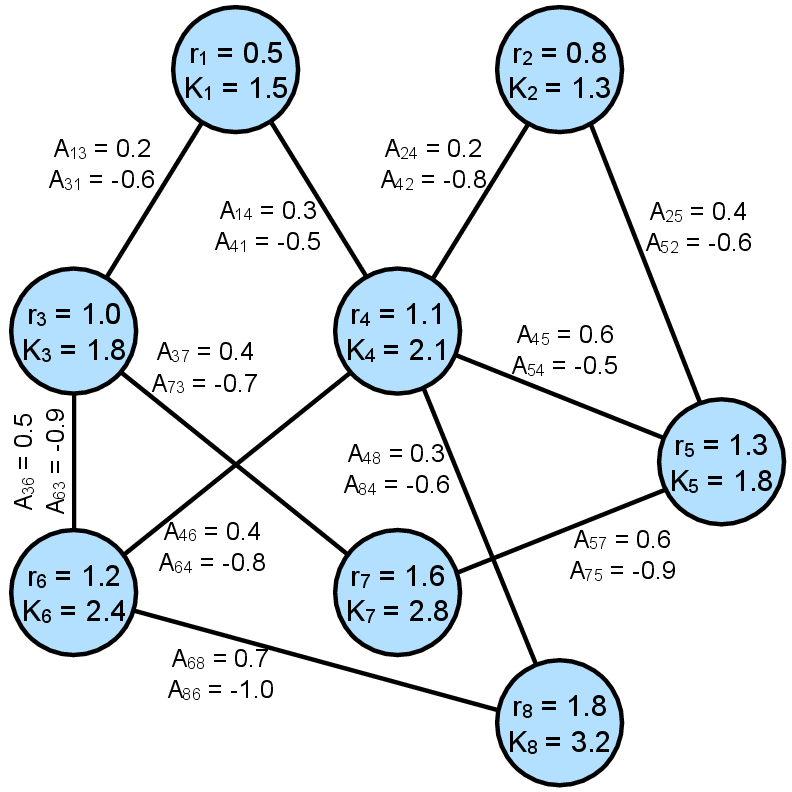, clip =,width=0.9\linewidth }
\caption{{\it Lotka-Volterra model.} Illustration of the $N=8$ species Lotka-Volterra model used here with the system given in Eq.~(\ref{eq:06}). All parameters (linear growth rates $r_i$, carrying capacities $K_i$, and interaction strengths $A_{ij}$) are given. Species are hierarchically organized with predators and prey at the top and bottom, respectively.}\label{fig2}
\end{figure}

\section{Example with Linear Forcing: Disturbances to a Food Web}\label{sec:03}

\begin{figure*}[t]
\centering
\epsfig{file =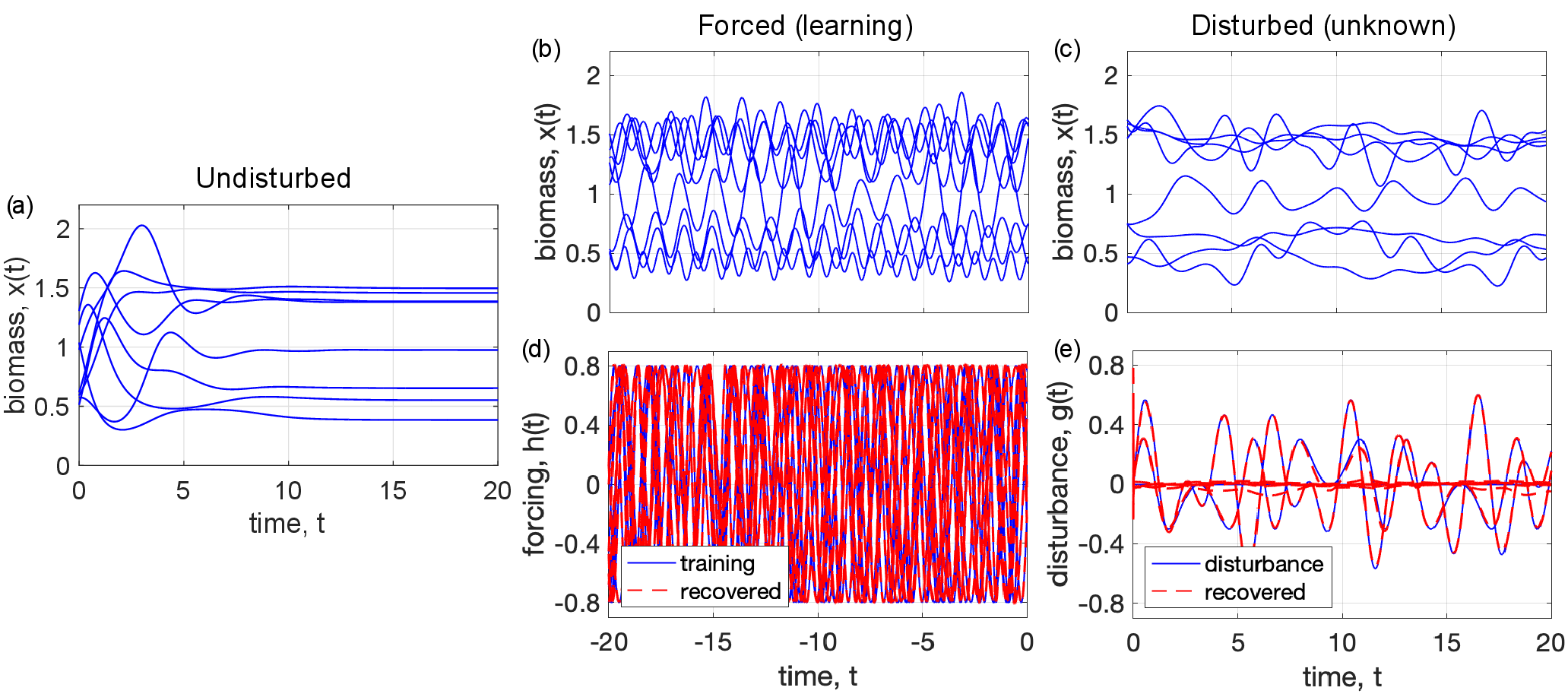, clip =,width=0.95\linewidth }
\caption{{\it Lotka-Volterra dynamics and disturbance recovery.} (a) Intrinsic dynamics of the undisturbed Lotka-Volterra system. (b),(c) biomasses $x_i(t)$ of the Lotka-Volterra system under sinusoidal forcing [$h_i(t)=0.8\sin(\omega_i t)$] and pseudo-sinusoidal disturbance [$g_3(t)=0.3\sin(2t)+0.3\sin(\pi t)$ and $g_5(t)=\sin(2\pi\sin(t/2))$], respectively. (d),(e) forcing and disturbance functions $h_i(t)$ and $g_i(t)$, respectively, with the actual and recovered functions plotted in solid blue and dashed red, respectively.}\label{fig3}
\end{figure*}

As a first example we consider the generalized Lotka-Volterra model, composed of $N$ interacting species whose populations (which we will also refer to as {\it biomasses}) are denoted $x_i(t)\ge0$. Note that the state of each species is a scalar, so each $x_i(t)$, $h_i(t)$, and $g_i(t)$ is of dimension $D=1$. Each biomass evolves according to
\begin{align}
\dot{x}_i = x_i\left(e_i - \frac{x_i}{K_i} + \sum_{j=1}^N P_{ij} x_j\right) + g_i(t),\label{eq:06}
\end{align}
where $e_i$ represents the linear growth rate of species $i$, $e_i K_i$ is the carrying capacity of species $i$, and $g_i(t)$ is the disturbance made to species $i$. Note that in this example disturbances are linear, i.e., each vector field $\bm{F}_i$ may be written as $\bm{F}_i(x_i,\bm{X},g_i(t))=\bm{F}_{i,\text{intrinsic}}(x_i,\bm{X})+g_i(t)$. The $N\times N$ matrix $P$ encodes the interactions between species such that $P_{ij}>0$ indicates that presence of species $j$ is favorable for species $i$, $P_{ij}<0$ indicates that presence of species $j$ is unfavorable for species $i$, and $P_{ij}=0$ if no direct interaction exists. Here we consider an $N=8$ species example whose interactions are those of predator-prey dynamics, i.e., for every interacting pair $(i,j)$, $P_{ij}$ and $P_{ji}$ are of opposite sign. In Fig.~\ref{fig2} we illustrate this network, indicating all parameters and organizing species hierarchically with predators at the top and prey at the bottom. The dynamics of the undisturbed system, i.e., with $g_i(t)=0$ for all $i=1,\dots,N$, are plotted in Fig.~\ref{fig3}(a) as the biomasses come to a stable fixed point representing coexistence. For this system we use a time step of $\Delta t=0.005$, updating the dynamics using Heun's method starting from a randomly chosen initial condition.

We now consider identifying and recovering unknown disturbances to this system using the reservoir computer architecture described above. For the training phase, we first consider a training interval of length $\hat{T}=100$ using forcing functions $h_i(t)$ composed of sinusoids with randomly chosen frequencies, specifically $h_i(t)=0.8\sin(\omega_i t)$, where each $\omega_i$ is randomly distributed in the interval $[1,9]$. We then consider disturbances made only to species $i=3$ and $5$ of the form $g_3(t)=0.3\sin(2t)+0.3\sin(\pi t)$ and $g_5(t)=\sin(2\pi\sin(t/2))$, and all other $g_i(t)=0$. Results are plotted in Fig.~(\ref{fig3}) with the biomasses $x_i(t)$ under the known forcing and the unknown disturbance plotted in panels (b) and (c), respectively, and the forcing functions and disturbance functions plotted in panels (d) and (e), respectively. The actual training and disturbances functions $h_i(t)$ and $g_i(t)$ are plotted in solid blue, while the recovered functions are plotted in dashed red. As we can see in Fig.~\ref{fig3}(e), despite the reservoir computer not knowing either the intrinsic system dynamics or the disturbances $g_i(t)$, the locations and nature of the disturbances are accurately recovered.

Before proceeding to a different example, a few remarks regarding the training forcing are in order. First, we used here a combination of sinusoids with mismatched (random) frequencies. This mismatch in frequencies allows (assuming a long enough training length $\hat{T}$) the forcing functions $h_i(t)$ to robustly explore all directions of forcing, which is necessary for accurately recovering disturbances~\cite{Canaday2021JPhys,RestrepoArxiv}. As we will see next, the form of the forcing given as sinusoids here can be generalized as long as a robust range of forcing is considered. Second, we also make note that in principle, some non-zero forcing should be used at each node in general. If a specific node is unforced during training, i.e., $h_i(t)=0$, then the method cannot recover any non-zero disturbance at that node.

\begin{figure}[t]
\centering
\epsfig{file =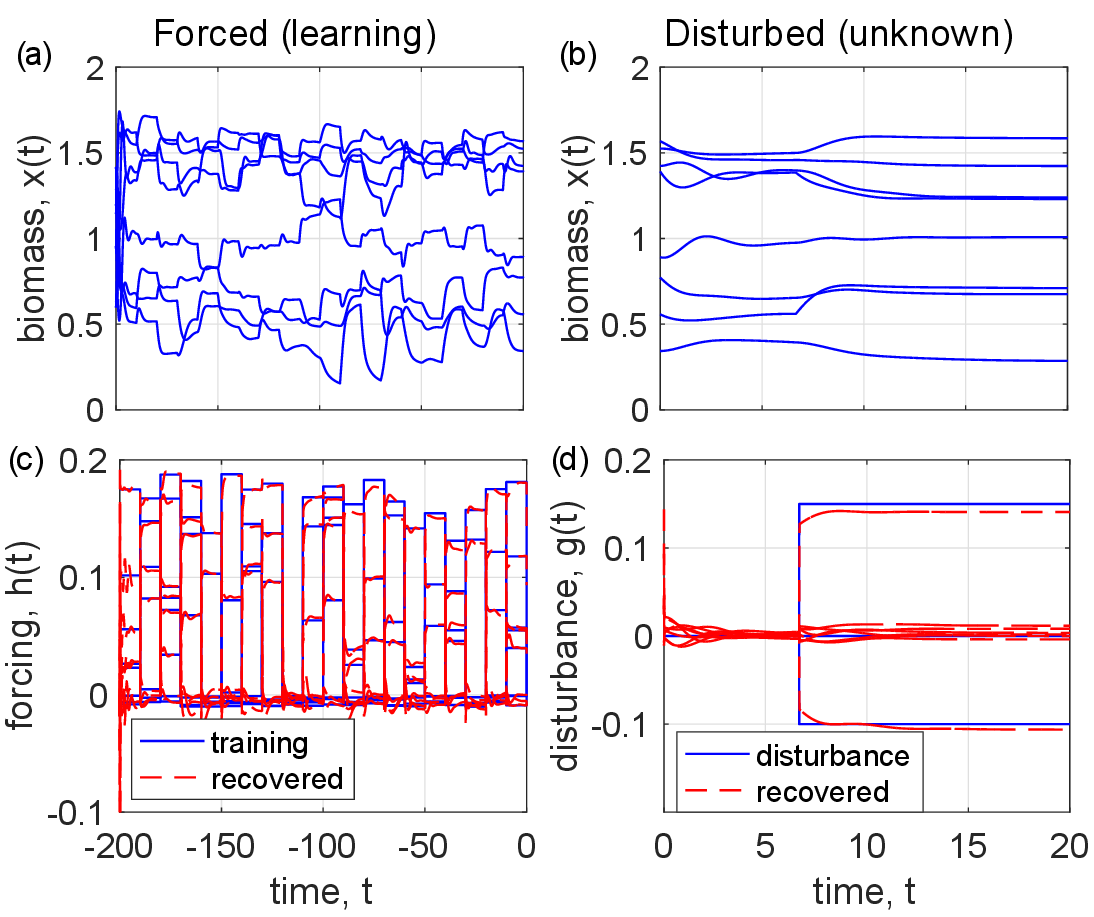, clip =,width=1.0\linewidth }
\caption{{\it Lotka-Volterra disturbance recovery: a second example.} (a),(b) biomasses $x_i(t)$ of the Lotka-Volterra system under random step forcing and Heaviside disturbance, respectively. (c),(d) forcing and disturbance functions $h_i(t)$ and $g_i(t)$, respectively, with the actual and recovered functions are plotted in solid blue and dashed red, respectively.}\label{fig4}
\end{figure}

In a second scenario, we consider a training interval of length $\hat{T}=200$ using forcing functions $h_i(t)$ composed of random step functions. Each $h_i(t)$ is partitioned into 20 intervals (each interval lasts $10$ time units) with values chosen uniformly at random in $[-0.01,0.19]$. Disturbances are then made again only to species $i=3$ and $5$ in the form of Heaviside functions that transition from $0$ to $-0.1$ and $0.15$, respectively, at $t=6.667$. Results are plotted in Fig.~(\ref{fig4}) with the biomasses $x_i(t)$ under the known forcing and the unknown disturbance plotted in panels (a) and (b), respectively, and the forcing functions and disturbance functions plotted in panels (c) and (d), respectively. Again, the actual training and disturbances functions $h_i(t)$ and $g_i(t)$ are plotted in solid blue, while the recovered functions are plotted in dashed red, showing a robust identification of the disturbances.

\section{Example with Nonlinear Forcing: Wilson-Cowan Neurons}\label{sec:04}

\begin{figure}[t]
\centering
\epsfig{file =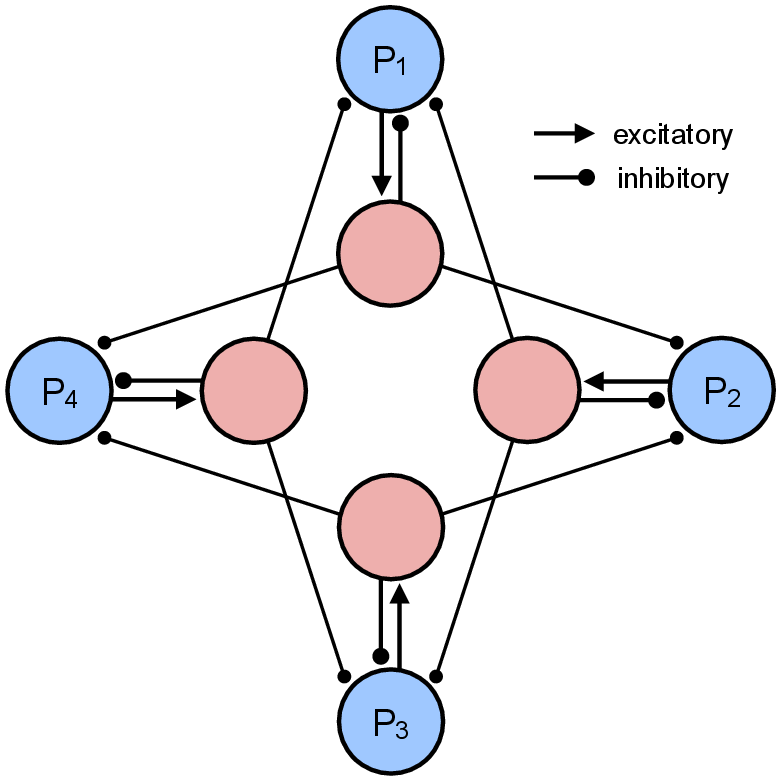, clip =,width=0.85\linewidth }
\caption{{\it Wilson-Cowan model.} Illustration of the $n=4$ ($N=8$) population Wilson-Cowan model used here with the system given in Eqs.~(\ref{eq:07})--(\ref{eq:09}). Inhibitory and excitatory populations are organized on the inside and outside, respectively.}\label{fig5}
\end{figure}

\begin{figure}[t]
\centering
\epsfig{file =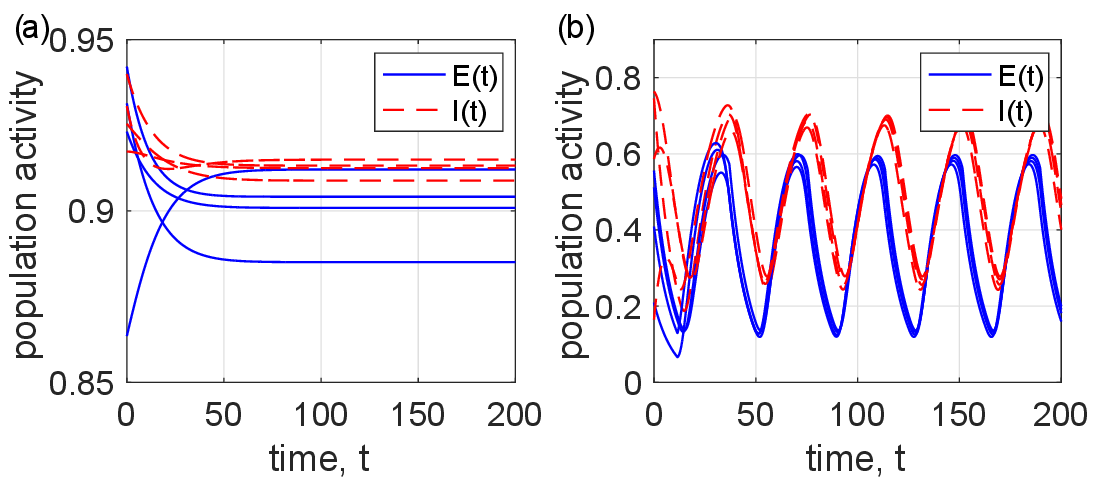, clip =,width=1.0\linewidth }
\caption{{\it Wilson-Cowan model: undisturbed dynamics.} Population activities $E_i(t)$ and $I_i(t)$ of the undisturbed Wilson-Cowan system in the (a) stationary and (b) oscillatory regimes.}\label{fig6}
\end{figure}

We now turn to a second example and consider a network of interacting Wilson-Cowan neuron populations. At its most minimal, the Wilson-Cowan model consists of one population of excitatory neurons and another population of inhibitory neurons whose aggregate activities are nonlinearly coupled, i.e., $D=2$. Here we consider a network consisting of $N=4$ pairs of populations (i.e., yielding a system of total size $ND=8$) where each pair consists of one excitatory population and one inhibitory population. The aggregate activity of each excitatory and inhibitory population are given by $E_i$ and $I_i$ with $i=1,\dots, N$ that evolve according to
\begin{align}
\tau\dot{E}_i &= -E_i + \nonumber\\
&S\left(w_{EE} E_i - w_{EI}I_i + P_i - w_{\text{net}}\sum_{j=1}^N B_{ij}I_j + g_i(t)\right)\label{eq:07}\\
\tau\dot{I}_i &= -I_i + S\left(w_{IE} E_i - w_{II}I_i\right),\label{eq:08}
\end{align}
where
\begin{align}
S(x) = \frac{Kx^2}{\sigma^2 + x^2},\label{eq:09}
\end{align}
is a sigmoidal response function. Note that each population has a linear response component in addition to the nonlinear response. Within the response function, both the excitatory and inhibitory populations play their namesake roles (coupling parameters $w_{EE}$, $w_{EI}$, $w_{IE}$, and $w_{II}$ are all positive), while the excitatory populations are fed a baseline stimulus $P_i$. Moreover, each pair is coupled through the adjacency matrix $B$ and network coupling parameter $w_{\text{net}}$ in such a way that inhibitory populations affect not only their local excitatory counterpart, but also other excitatory populations according to the adjacency matrix. Finally, note that the disturbances in this system are nonlinear, i.e., functions $g_i(t)$ appear within the sigmoidal response functions, leading to nonlinear disturbances. Our system of $N=4$ pairs of populations is illustrated in Fig~\ref{fig5}. Parameters are given by $\tau = 10$, $w_{EE}=6.4$, $w_{EI} = 6.0$, $w_{IE}=4.8$, $w_{II}=1.2$, $w_{\text{net}}=0.5$, $K=1$, and $\sigma=1$, as well as two sets of baseline stimuli $P_i$ to give rise to two different dynamical regimes. Specifically, we consider a larger set of stimuli, $\bm{P}=[3.22,3.02,3.18,3.33]^T$ and a set of smaller stimuli, $\bm{P}=[1.52,1.61,1.57,1.64]^T$, which give rise to stationary dynamics and oscillatory dynamics, respectively, as illustrated in Fig.~\ref{fig6}(a) and (b). Here we use a time step of $\Delta t=0.2$. We also consider disturbances made only to the excitatory nodes, which can be interpreted as modifications of the external stimuli $P_i$. Thus, in the notation adopted in the problem description, each state vector is given by $\bm{x}_i(t)=[E_i(t),I_i(t)]^T$ and we consider forcing and disturbances of the form $\bm{h}_i(t)=[h_i(t),0]^T$ and $\bm{g}_i(t)=[g_i(t),0]^T$.

\begin{figure}[t]
\centering
\epsfig{file =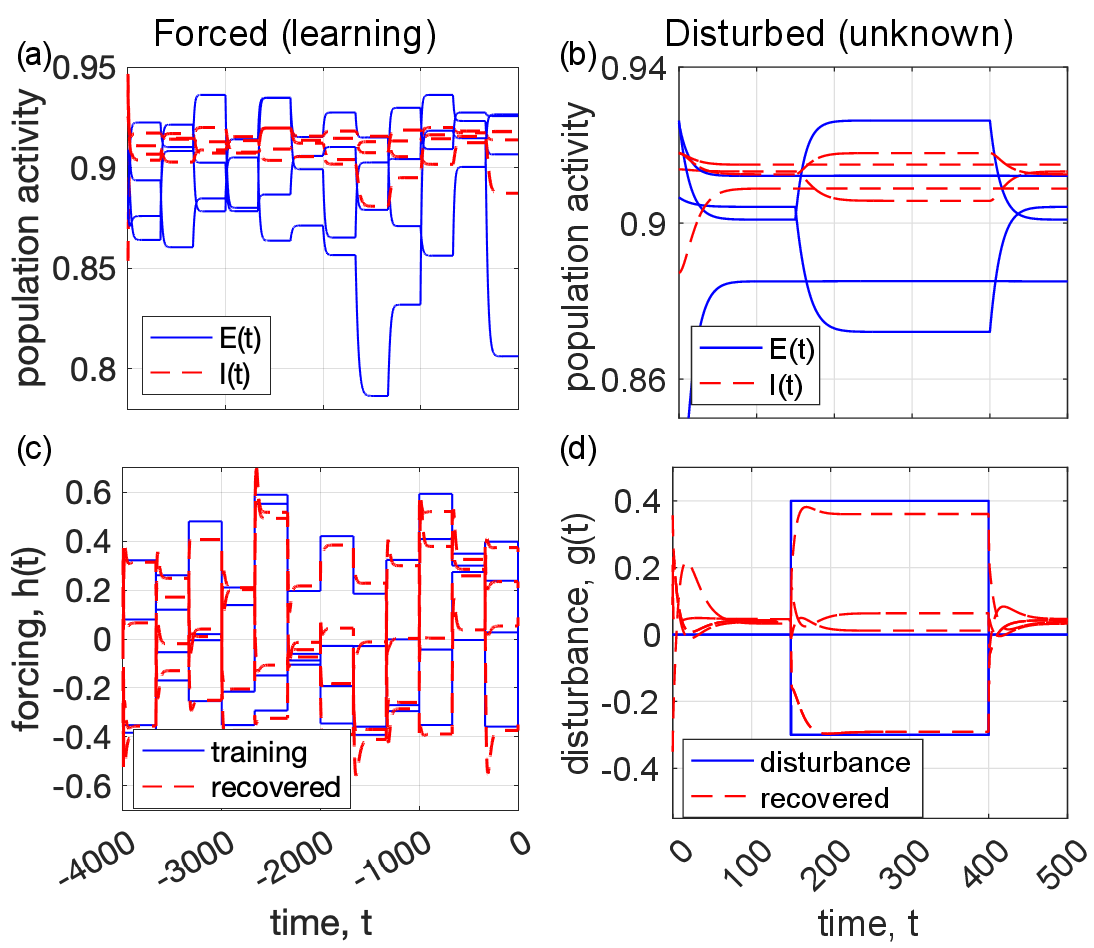, clip =,width=1.0\linewidth }
\caption{{\it Wilson-Cowan disturbance recovery: stationary regime.} (a),(b) Population activities $E_i(t)$ and $I_i(t)$ of the Wilson-Cowan system under random step forcing and Heaviside disturbance, respectively. (c),(d) forcing and disturbance functions $h_i(t)$ and $g_i(t)$, respectively, with the actual and recovered functions are plotted in solid blue and dashed red, respectively.}\label{fig7}
\end{figure}

\begin{figure*}[t]
\centering
\epsfig{file =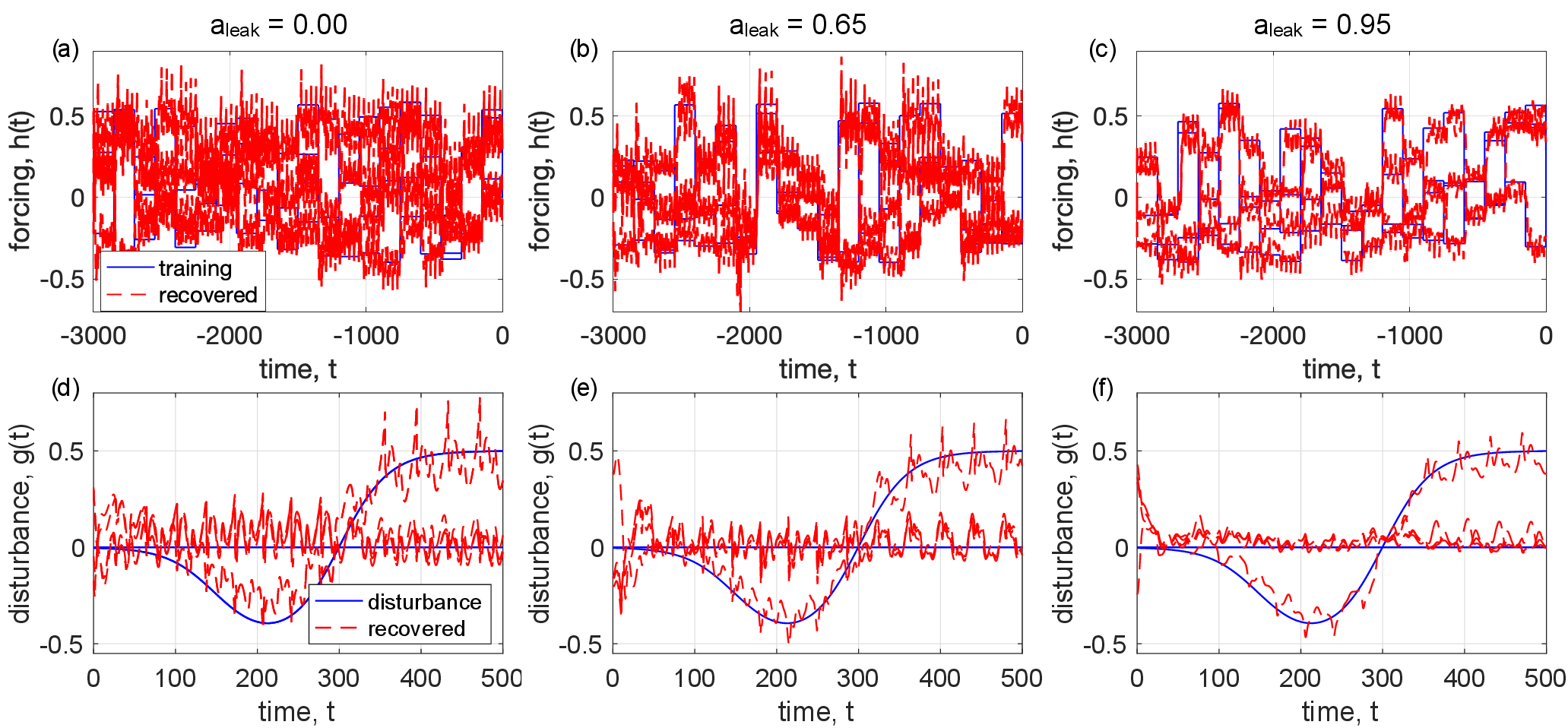, clip =,width=0.95\linewidth }
\caption{{\it Wilson-Cowan disturbance recovery: oscillatory regime.} Using leak parameters $a_{\text{leak}}=0$, $0.65$, and $0.95$ the actual and recovered (solid blue and dashed red, respectively) forcing functions $h_i(t)$ plot in panels (a), (b), and (c), along with corresponding actual and recovered disturbance functions $g_i(t)$ plotted in panels (d), (e), and (f).}\label{fig8}
\end{figure*}

Beginning with the dynamics in the stationary regime (i.e., we use  $\bm{P}=[3.22,3.02,3.18,3.33]^T$), we consider a training interval of length $\hat{T}=4000$ and, drawing from the previous example, use forcing functions $h_i(t)$ composed of random steps. Each $h_i(t)$ is partitioned into 12 intervals with values chosen uniformly at random in $[-0.4,0.6]$. Disturbances are then made again only to populations $i=1$ and $3$ in the form of Heaviside functions that transition from $0$ to $0.4$ and $-0.3$, respectively, at $t=150$ and are back to zero at time $t=400$. Results are plotted in Fig.~\ref{fig7} with the population activity $E_i(t)$ (solid blue) and $I_i(t)$ (dashed red) under the known forcing and the unknown disturbance plotted in panels (a) and (b), respectively, and the forcing functions and disturbance functions plotted in panels (c) and (d), respectively. Again, the actual training and disturbances functions $h_i(t)$ and $g_i(t)$ are plotted in solid blue, while the recovered functions are plotted in dashed red, showing a robust identification of the disturbances.

Moving on to the oscillatory regime (i.e., using $\bm{P}=[1.52,1.61,1.57,1.64]^T$), we find more nuanced results. Maintaining the same training setup as above, we now consider disturbing a single excitatory population with a composed sigmoidal function, $g_1(t)=-\frac{1}{2}\frac{1}{1+\text{exp}(-(t-150)/30)} + \frac{1}{1+\text{exp}(-(t-300)/30)}$, which effectively first decreases, then increases the baseline stimulus. Our initial attempt at recovering this disturbance is convoluted by the oscillatory dynamics, as can be seen in the results plotted in Figs.~\ref{fig8}(a) and (d), where we plot the known forcing and the unknown disturbance, respectively, again with solid blue and dashed red denoting the actual and recovered signals. Note that the reservoir has trouble recovering the training functions $h_i(t)$, specifically averaging out the system's oscillations, which propagates into the recovery of the disturbances $g_i(t)$. Overall, the reservoir computer is able to produce a signal whole temporally-localized mean is accurate, but losing precision due to strong oscillations. To address this issue we adjust the reservoir dynamics to include a leak parameter $a_{\text{leak}}\in[0,1]$, specifically, we consider dynamics of the form
\begin{align}
\bm{r}(t+\Delta t) &= a_{\text{leak}}\bm{r}(t) \nonumber\\
&+ (1-a_{\text{leak}})\tanh[A \bm{r}(t) +W_{\text{in}} \hat{\bm{x}}(t) + 1],\label{eq:07}
\end{align}
where $a_{\text{leak}}=0$ recovers the reservoir dynamics used previously [i.e., Eq.~(\ref{eq:03})], and with larger values essentially slowing down the dynamics. Using this adjustment of the reservoir dynamics, first plotting the forcing and disturbance results in Figs.~\ref{fig8}(b) and (e) for $a_{\text{leak}}=0.65$, then in Figs.~\ref{fig8}(c) and (f) for $a_{\text{leak}}=0.95$, quenches the oscillations recovered in both the training and disturbance phases, improving results. This improvement, however, comes at the cost of slowing down the reservoir dynamics and choosing $a_{\text{leak}}$ too large can eventually degrade results, depending on the dynamics and forcing/disturbance.

\section{Scalability to Large Networks}\label{sec:05}

Before closing we consider the case of larger networks than those presented above, and how this framework can be scaled to obtain accurate results, even for networks with many more units. For this purpose, we return to the Lotka-Volterra model used in Sec.~\ref{sec:03} which describes biomasses $x_i(t)$ for $i=1,\dots,N$ via Eq.~(\ref{eq:06}). To generate such systems for arbitrary size $N$ we consider networks where each species has on average four interactions, a minimum of two interactions, and each non-zero interaction strength $P_{ij}$ has absolute value randomly and uniformly drawn from the interval $[2/N,4/N]$ with all upper-triangular entries ($j>i$) being positive and lower-triangular entries ($j<i$) positive (negative) with probability $1/2$ ($1/2$). Note that this suggests that roughly half of all interactions represent a predator-prey interaction while the other half are mutualistic. In addition, each growth rate $e_i$ is drawn randomly and uniformly from the interval $[2,4]$ and each $K_i$ is drawn randomly and uniformly from the interval $[1,2]$. We also ensure that a strictly positive equilibrium of coexistence exists, with all species reaching equilibria at a value as large or larger than one, i.e., all fixed point values $x_i^*$ satisfy $x_i^*\ge1$.

\begin{figure}[t]
\centering
\epsfig{file =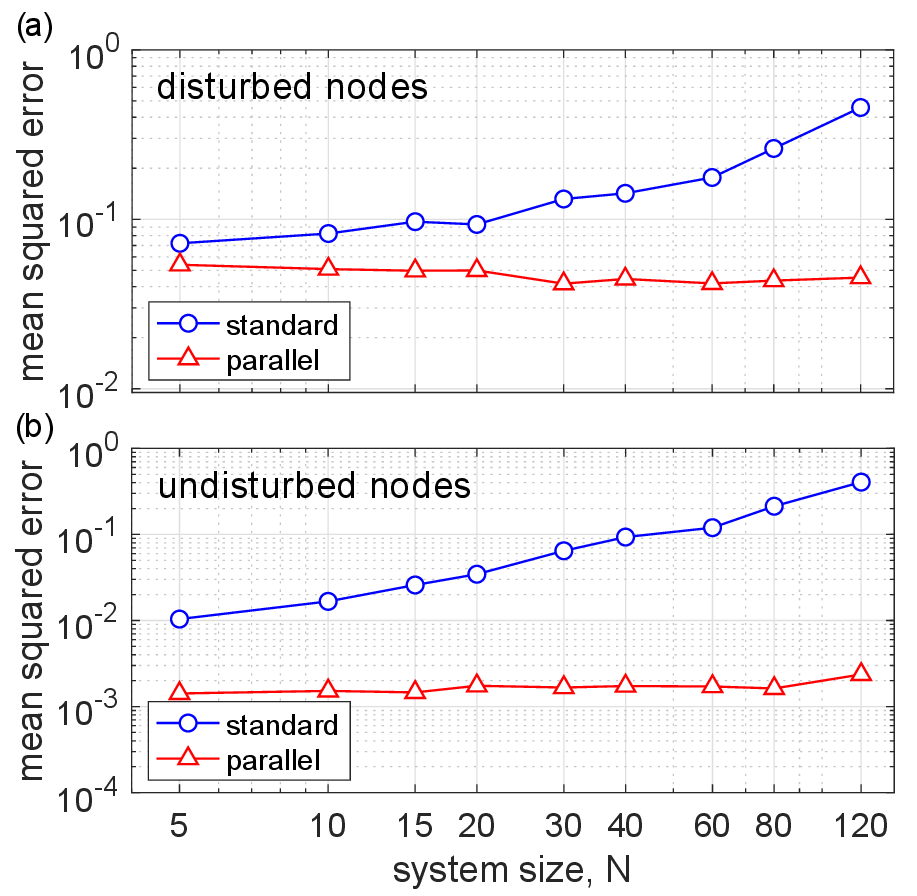, clip =,width=0.95\linewidth }
\caption{{\it Standard vs pseudo-parallel schemes: Error comparison.} As a function of system size $N$, a comparison of the mean squared error for the standard (blue circles) and pseudo-parallel (red triangles) schemes that use, respectively, one large reservoir of size $NM$ and $N$ smaller reservoirs or size $M$, with $M=100$. Results in panels (a) and (b) represent error calculated from disturbed and undisturbed nodes, respectively.}\label{fig9}
\end{figure}

Designing a reservoir computer that perform accurately as system size increases, however, is nontrivial. The predictive ability of traditional reservoir computer implementations quickly degrades as the network size grows~\cite{SrinivasanPRL2022} even as the size of the reservoir scales with the size of the system. While several methods have been recently explored and shown to improve the performance of reservoir computers~\cite{Carroll2019Chaos,Shirin2019Chaos,DelFrate2021Chaos,Hart2023Chaos,Nathe2023Chaos}, here we follow the architecture presented in Ref.~\cite{SrinivasanPRL2022}. Specifically, we consider an alternative where $N$ smaller reservoirs are used: one for each unit $i$ that takes as an input the value $x_i(t)$ and the values of the biomasses $x_j(t)$ of its corresponding network-neighbors. For a system with $N$ dynamical units we then consider two different reservoir architectures: (i) a standard single reservoir of size $NM$ and (ii) a pseudo-parallelized architecture with $N$ reservoirs (one for each unit) each of size $M$. Here the natural number $M$ is a parameter that defines the size of the reservoirs in the pseudo-parallelized scheme and scales the size of the reservoir in the standard scheme to ensure that the two methods have the same total number of reservoir nodes.

We now compare the results obtained using the standard and pseudo-parallelized architectures for Lotka-Volterra systems described above over a range of system sizes $N$ with reservoir size parameter $M=100$. For each case we use a training interval of length $\hat{T}=100$ with training functions $h_i(t)=0.8\sin(\omega_i t)$, where each frequency $\omega_i$ is randomly and uniformly drawn from the interval $[1,9]$. We then consider disturbances to $20\%$ of nodes (the remaining $80\%$ are undisturbed) with either the function $g_i(t) = u_1\sin(u_2 t) + u_1\sin(u_3 t)$ or $u_1\sin[u_4\sin(u_2 t)]$ (chosen with probabilities $1/2$ and $1/2$) where, for each node, $u_1$, $u_2$, $u_3$, and $u_4$ are randomly and uniformly drawn from the intervals $[0.2,0.4]$, $[1,2]$, $[\pi/2,\pi]$, and $[\pi,2\pi]$, respectively. We consider $20$ realization of each system size $N$, plotting in Fig.~\ref{fig9} the mean squared error aggregated over all realizations for the standard and pseudo-parallel schemes in blue circles and red triangles, respectively. Results plotted in panels (a) and (b) represent errors taken from disturbed and undisturbed nodes, respectively. As system sizes increases from $N=5$ to $120$ we see that while the error in results obtained by using the standard reservoir scheme increase steadily, when the pseudo-parallelized scheme is used the error remains roughly constant. Thus, by implementing the pseudo-parallelized scheme, the overall framework for identifying disturbances in network-coupled dynamical systems scales and remains robust even as larger networks are considered, maintaining accurate results while providing significant computational savings for large $N$.

\section{Discussion}\label{sec:06}

In this work we have presented a model-free method for identifying disturbances in networks of coupled dynamical systems. This method is based on the machine learning framework of reservoir computing. No knowledge of either the underlying dynamics or the nature of the disturbance itself is assumed. In fact, all that is required is the observed behavior of the system under a known training forcing function.

Using food web and neuronal population models as examples, we demonstrated that this method robustly identifies disturbances to network-coupled dynamical systems using a range of relatively simple forcing functions for training. For certain cases where the dynamics are periodic we have shown that including a leak in the reservoir dynamics can improve results. Moreover, by implementing a pseudo-parallel reservoir architecture we illustrated that this method robustly scales with system size. This parallelization maintains accurate results while providing efficient computation for large networks.

As machine learning techniques become more and more integrated into dynamical systems theory, we emphasize that the technique presented here differs from most applications of reservoir computing to nonlinear dynamics in that our aim is not to learn the intrinsic dynamics of a system, but rather to infer some some extrinsic information, here an external disturbance. We believe that similar techniques may be useful for a wider variety of tasks that go beyond forecasting system behaviors, for example inferring model parameters, network structures, or the other properties of interactions, e.g., coupling functions.

\acknowledgments
PSS acknowledges support from NSF grant MCB-2126177. JGR acknowledges support from NSF Grant DMS-2205967.

\section*{Author Declarations}
\subsection*{Conflict of Interest}
The authors have no conflicts to disclose.

\bibliographystyle{unsrt}

\end{document}